%% file: ICPR_2026_LaTeX_Templates/main.tex
\documentclass[runningheads]{llncs}
\usepackage[T1]{fontenc}
\usepackage{amsmath} 
\usepackage{graphicx}
\usepackage{multirow}
\usepackage{booktabs}
\usepackage{pifont}
\newcommand\parahead[1]{\vspace{1.5mm}\noindent\textbf{#1.}\medspace}

\input{ICPR_2026_LaTeX_Templates/preamble}

\begin{document}
\title{Gaze-Anchored Social Net: \\Decoding Implicit Relations via Joint Modeling}
%
%
\author{Yuqi Hou\inst{1}\orcidID{0009-0002-3086-8698} \and
Zhuo Chen\inst{1}\orcidID{0000-0003-2988-9387} \and
Han Hu\inst{1}\orcidID{0000-0002-9667-1698} \and
Je Woo Kim\inst{2}\orcidID{0009-0004-5073-7307} \and
Jianbo Jiao\inst{1}\orcidID{0000-0003-0833-5115} \and
Hyung Jin Chang\inst{1}\orcidID{0000-0001-7495-9677}}
\authorrunning{Y. Hou et al.}
%
\institute{University of Birmingham, United Kingdom \newline
\email{\{yxh029,zxc417,hxh347\}@student.bham.ac.uk, \{h.j.chang,j.jiao\}@bham.ac.uk}
\\
\and
Korea Electronics Technology Institute, Korea\\
\email{jwkim@keti.re.kr}}
\maketitle      

\input{sec/0_abstract}    
\input{sec/1_intro}
\input{sec/2_relatedwork}
\input{sec/3_dataset}
\input{sec/4_method}
\input{sec/5_exp}

\input{sec/6_discussion}
\input{sec/7_conclusion}

\bibliographystyle{splncs04} 
\bibliography{sec/refs}  

\end{document}

%% file: ICPR_2026_LaTeX_Templates/preamble.tex
\usepackage{makecell}
\usepackage{comment}
\usepackage{multirow}
\usepackage{suffix}
\usepackage[dvipsnames]{xcolor}

\newcommand\authorcomment[3]{\noindent\textsf{\textcolor{#1}{[\textbf{#2:} \textit{#3}]}}}
\WithSuffix\newcommand\authorcomment*[2]{\noindent\textcolor{#1}{\textit{#2}}}

\newcommand{\yuqi}[1]{\authorcomment{PineGreen}{Yuqi}{#1}}
\WithSuffix\newcommand\yuqi*[1]{\authorcomment*{PineGreen}{#1}}

\newcommand{\yh}[1]{\authorcomment{BurntOrange}{Yihua}{#1}}
\WithSuffix\newcommand\yh*[1]{\authorcomment*{BurntOrange}{#1}}

\newcommand{\swook}[1]{\authorcomment{Maroon}{Wookie}{#1}}
\WithSuffix\newcommand\swook*[1]{\authorcomment*{Maroon}{#1}}

\newcommand{\hfw}[1]{\authorcomment{NavyBlue}{Hengfei}{#1}}
\WithSuffix\newcommand\hfw*[1]{\authorcomment*{NavyBlue}{#1}}

\newcommand{\hjc}[1]{\authorcomment{Fuchsia}{Hyung}{#1}}
\WithSuffix\newcommand\hjc*[1]{\authorcomment*{Fuchsia}{#1}}

\newcommand{\chzh}[1]{\authorcomment{Violet}{Zhuo}{#1}}
\WithSuffix\newcommand\chzh*[1]{\authorcomment*{Voilet}{#1}}

\WithSuffix\newcommand\parahead*[1]{\vspace{2mm}\noindent\textbf{#1}\medspace}

\usepackage{pifont}
\newcommand{\cmark}{\color{ForestGreen} \ding{51}}%
\newcommand{\xmark}{\color{Red} \ding{55}}%

\usepackage{amsmath}

\usepackage{svg}

\usepackage{graphicx}
\usepackage{booktabs}
\usepackage{subcaption} 
\usepackage{makecell}  
\usepackage{amssymb}

%% file: sec/0_abstract.tex
\begin{abstract}
Human gaze does more than point to visual targets; it serves as a subtle indicator of social intent within static images, whereas standard models typically process individuals independently, treating gaze as an $\text{i.i.d.}$ quantity or predicting social semantics in isolation. 
Recent multi-person methods attempt to address this but often treat social relations as rigid, post-hoc classifications decoupled from the gaze estimation process. 
This oversimplification fails to capture the nuanced nature of social intent, which acts as an underlying driver of gaze behavior rather than a secondary categorical output.
We address these limitations by proposing ANCHOR, a target-centric paradigm designed to decode gaze-anchored social intent by modeling the joint distribution of visual attention and latent implicit relations.
Our approach surfaces these dependencies as the latent structural scaffolding of gaze behavior. The architecture utilizes a relational attention mechanism to capture fine-grained interpersonal links, leveraging feature-wise modulation for efficient multi-person parsing from a single vision backbone.
To stabilize the training of this coupled formulation, we implement an optimization synergy to resolve the inherent conflicts between spatial gaze accuracy and latent social reasoning. This approach ensures robust generalization by seeking stable, flat minima while simultaneously harmonizing competing task gradients. 
We validate our framework on an extended benchmark featuring dense multi-person annotations and novel social influence rankings.
Our results demonstrate state-of-the-art performance and provide the first quantitative evidence that implicit social hierarchies can be robustly disentangled and learned directly from static gaze patterns.

\keywords{social net  \and static multi-person gaze \and optimization synergy.}
\end{abstract}

%% file: sec/1_intro.tex
\section{Introduction}
\label{sec:intro}

Human gaze conveys a wide range of social cues. Prior work shows that where people look reflects their mental states, interactions, and these signals extend well beyond the role of gaze in basic visual processing~\cite{Mayrand2024,mayrand2024gaze,Ristic2022}.
In multi-person scenes, gaze patterns encode rich information about social intent, and interpersonal relationships that extend far beyond individual visual attention.

Despite this potential, current gaze-following methods~\cite{hou2024multi,ryan2025gaze} share a key limitation: 
they process individuals independently, modeling gaze as $P(X_{\text{gaze}}^{(i)} \mid X_{\text{scene}})$ for each person, implicitly assuming conditional independence across individuals and ignoring inter-person interactions.

Even multi-person models~\cite{tafasca2024sharingan,ryan2025gaze} primarily optimize for efficiency while maintaining this independence assumption. 
Other approaches~\cite{gupta2024unified,gupta2025mtgs} add auxiliary social tasks such as looking-at-head, shared attention, or looking-at-each-other, 
but these remain pairwise and rely on coarse, discrete labels within a decoupled framework that models gaze and social cues separately as  $P(X_{\text{gaze}} \mid X_{\text{scene}})$ and $P(X_{\text{social}} \mid X_{\text{gaze}}, X_{\text{scene}})$, which cannot express the continuous subtleties of social behavior.
Similar limitations appear in broader social group research~\cite{jahangard2024jrdb}.

We argue that the challenge is not to infer social intent post-hoc as  $P(X_{social} | \allowbreak X_{gaze}, \allowbreak X_{scene})$, but rather to model the joint distribution $P(X_{gaze}, X_{social} | X_{scene})$. This formulation reveals a crucial observation: features necessary for accurate gaze prediction are inherently entangled with latent social intent $X_{social}$. In other words, a model achieving high accuracy on gaze prediction must have already implicitly learned to decode social dynamics to succeed. The real challenge, therefore, is not adding discrete labels, but explicitly surfacing this latent information already encoded in the learned features.

Surfacing this latent social information introduces three fundamental challenges that have not been addressed in prior work:
(i) latent decoding of implicit social intent without explicit supervision; (ii) synergistic optimization to mitigate gradient interference between gaze and social tasks; and (iii) quantitative validation in the absence of established benchmarks.

Preliminary experiments indicated that modeling global social networks via GNNs introduces significant noise from distant agents, while explicit social annotations often prove subjective and unreliable. 
Social perception is naturally organized from a self-centered reference point, where observers prioritize cues relative to a focal individual~\cite{myers2020ebook}.
Leveraging the relational richness inherent in modern vision backbones~\cite{oquab2023dinov2}, we propose a target-centric paradigm. Instead of a noisy global graph, we decode social dynamics through the subjective perspective of the target head, filtering spatial noise while preserving high-fidelity relational context.

To address the Decoding Challenge, we introduce the ANCHOR(Fig~\ref{fig:framework}) framework, designed to surface the implicit relations that drive visual attention. At its core, the architecture employs a relational attention mechanism that models the local interplay between a target individual and their geometric neighbors. By treating the target’s features as a query against its surroundings, the model explicitly decodes the joint dependencies that constitute gaze-anchored social net. This relational context is then integrated through feature-wise modulation, allowing for precise, social-aware parsing of multiple individuals from a single frozen encoder.

Jointly optimizing gaze and social intent triggers gradient conflicts that destabilize training. We resolve these by enforcing flat minima and harmonizing gradients to mitigate interference. A self-supervised consistency objective provides geometric signals to refine social modules without manual labels.

\input{fig_tex/framework}

To address the Validation Challenge, we introduce two novel extensions to the GazeFollow test set: (1) multi-person gaze annotations to validate joint-distributional modeling, and (2) Social Influence Ranking (SIR) labels. By capturing collective gaze convergence, SIR transforms ambiguous social perceptions into a measurable ranking task. We evaluate this via the Social Consistency Index (SCI) to validate internal geometric mechanisms and Social Leader Accuracy (SLA) to measure practical leader recognition. Together, these benchmarks provide the first quantitative evidence that implicit social dynamics can be robustly disentangled and learned.

Our work is the first to identify and address the fundamental limitations in current gaze-social modeling paradigms. Our specific contributions are:
\begin{itemize}
    \item We propose ANCHOR, a target-centric framework that decodes a gaze-anchored social net through relational attention and a self-supervised $L_{SCI}$ loss, uncovering latent social dynamics without manual supervision.
    \item Our synergistic optimization strategy resolves gradient conflicts in multi-task learning, ensuring stable convergence and robust generalization.
    \item We introduce the Extended GazeFollow benchmark with multi-person gaze and Social Influence Ranking (SIR) labels, achieving state-of-the-art multi-person gaze prediction performance and providing the first quantitative validation of learned social dynamics.
\end{itemize}

%% file: fig_tex/framework.tex
\begin{figure*}[t]
    \centering
    \includegraphics[width=\linewidth]{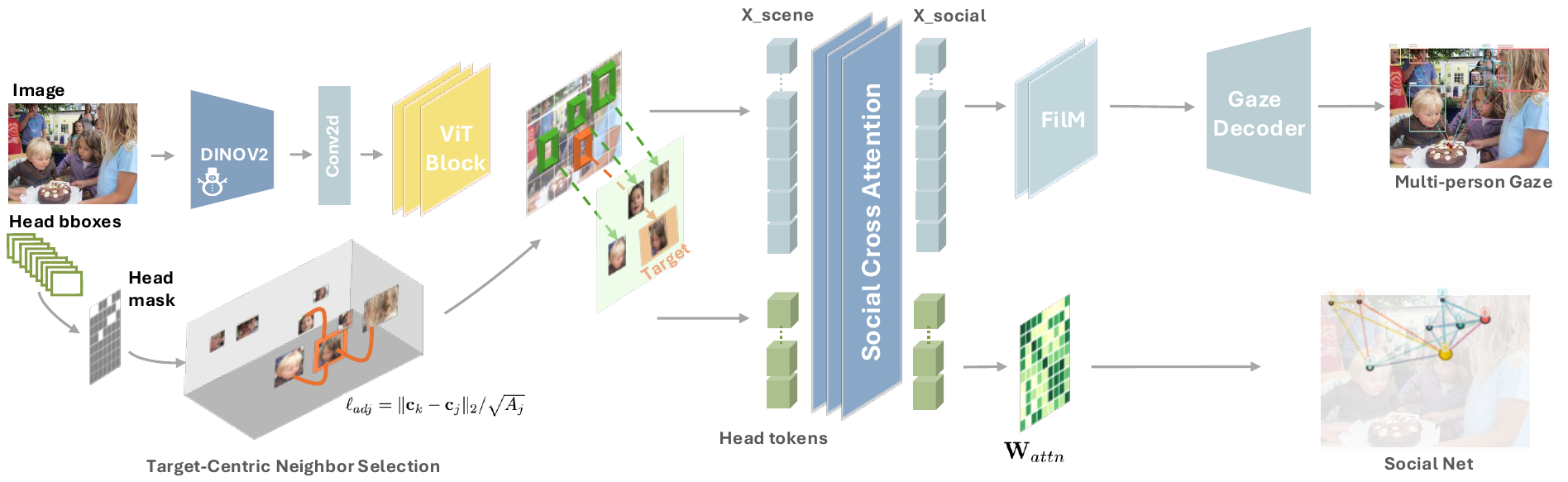}
    \caption{
        Overview of the ANCHOR Framework. 
    }
    \label{fig:framework}
\end{figure*}

%% file: sec/2_relatedwork.tex
\section{Related Work}
\label{sec:related}
\parahead{From Independent to Multi-person Gaze Following}
Early gaze-following methods typically relied on a dual-stream architecture integrating head-orientation cues with spatial saliency maps~\cite{recasens2015they,recasens2016following,sumer2020attention}. 
Subsequent research has progressively enhanced this foundation by incorporating diverse modalities such as 3D gaze models~\cite{chong2018connecting}, scene depth~\cite{fang2021dual,hu2022we,horanyi2023they}, and panoramic context~\cite{Li_2021_ICCV}, or leveraging high-level semantics including human-object interactions~\cite{wei2018looking,tonini2023object} and audio-visual cues~\cite{hou2024multi}.
Despite providing robust features, these works primarily target single-subject scenarios. Recent multi-person methods~\cite{tafasca2024sharingan,gupta2024unified,gupta2025mtgs} add auxiliary social tasks but remain constrained by coarse, discrete interaction categories.

\parahead{The Conditional Independence Fallacy}
Recent architectures treat multi-person gaze as a set-prediction problem, yet often rely on a conditional independence fallacy. For instance, several frameworks~\cite{tu2022end,ryan2025gaze,tafasca2024sharingan} process individual heads in isolation or suffer from noise-prone padding due to rigid architectural constraints. Even methods that incorporate auxiliary social tasks~\cite{gupta2024unified,gupta2025mtgs} remain limited to coarse, pairwise categories, failing to capture the continuous, joint-distributional nature of social behavior.
In contrast, ANCHOR explicitly decodes joint dependencies by modeling $P(X_{gaze}, X_{social} | X_{scene})$ through a flexible, padding-free framework. 

\parahead{Social Interaction and Group Detection}
Traditional research identifies explicit social groups via discrete labels, ranging from early F-formations~\cite{setti2015f,alameda2015salsa} to categorical IDs in recent datasets like JRDB-Social~\cite{jahangard2024jrdb}. While effective for high-level classification, these oversimplified, binary labels suffer from subjective annotation bias and fail to capture the continuous subtleties of implicit social intent. ANCHOR diverges by rejecting discrete labels to decode latent social dynamics directly from gaze features. This paradigm is validated through our novel Social Influence Ranking (SIR) metric, which transforms subjective social perceptions into a continuous, quantifiable measure of intent.

\parahead{Gaze Following Datasets}
GazeFollow~\cite{recasens2015they} established the foundation for large-scale gaze detection, with subsequent datasets addressing temporal dynamics (VAT~\cite{chong2020detecting}), shared attention (VideoCoAtt~\cite{fan2018inferring}), object context (VACATION~\cite{fan2019understanding}, GOO~\cite{tomas2021goo}), and specialized domains~\cite{tafasca2023childplay,hou2024multi}. However, these lack the comprehensive multi-person annotations and quantitative ranking labels necessary for joint-distributional modeling. Existing multi-person works often rely on subjective binary labels (e.g. LAH, SA, LAEO), which fail to capture the nuanced spectrum of social influence. To bridge this gap, our Extended GazeFollow benchmark introduces: (1) multi-person gaze annotations and (2) Social Influence Ranking (SIR) labels. This allows us to move beyond discrete classification toward a fine-grained, quantitative understanding of implicit social dynamics.

%% file: sec/3_dataset.tex
\section{Extended GazeFollow Benchmark}
While existing datasets lack the quantitative social measures required for decoding (Fig.~\ref{tab:datasets}) , the original GazeFollow~\cite{recasens2015they} offers a uniquely diverse range of multi-person social scenarios (Fig.~\ref{fig:heads_distribution}). To bridge this gap, we enriched its 3,018 test images with two novel layers of manual annotations.

\parahead{Multi-Head Gaze Annotations}
To evaluate true multi-head performance, we provide comprehensive gaze annotations for \textit{all} valid heads in all 3,018 test images, using refined head boxes from~\cite{tafasca2024sharingan}. We deprecated heads with minimal information ($area < 4~\text{px}^2$). This resulted in 17,313 total labels, providing dense supervision across complex social scenes (36\% of images contain $>3$ heads). As shown in Fig.~\ref{fig:head_number_hist}, our benchmark provides over 86\% label coverage for images with many heads, correcting a key deficiency in the original sparse dataset.

\parahead{Social Influence Ranking (SIR) Labels}
To create a ground truth for the latent variable $X_{social}$, we introduce SIR labels. We engaged 4 human annotators to rank all surrounding individuals based on their perceived social influence on the primary target. This ordinal ranking (Top-15) was successfully applied to all 1,428 test images containing sufficient social context ($>= 2$ neighbors).
\input{tab/comparison_gazefollow_multi}

\parahead{Annotation Procedures and Annotator Agreement}
Since GazeFollow typically annotates only the most salient head in each image, we treat this original head as the target subject. Based on this target subject, we first add multi-head gaze annotations: we remove incorrectly detected, overly small, or visually unclear head boxes, and then annotate gaze points for each valid head using the same protocol as the original dataset. We also use the same target subject for Social Influence Ranking (SIR), using the interface shown in Fig.~\ref{fig:social_impact_UI}. Annotators assign an A–O influence level (ranging from highest to lowest social impact) to every other person in the image using only visual cues such as body orientation, depth-adjusted distance, and simple signs of interaction. Each person receives a unique rank, and no gaze labels or geometric metadata are provided. All four annotators follow the same instructions, and although small differences appear in fine-grained cases, the overall patterns are consistent. We merge their labels into a single consensus ranking, which serves as the final ground truth.



%% file: tab/comparison_gazefollow_multi.tex
\begin{table}[t]
    \centering
    \begin{minipage}[c]{0.58\linewidth}
        \centering
        \captionof{table}{\textbf{GazeFollow Benchmark Comparison.}}
        \label{tab:gazefollow_multi}
        \resizebox{\linewidth}{!}{%
            \begin{tabular}{l | c c c | c c c}
            \toprule
            \multirow{2}{*}{\textbf{Method}} &
            \multicolumn{3}{c|}{\textbf{Multi-person Gaze}} &
            \multicolumn{3}{c}{\textbf{Single-person Gaze}} \\
            \cmidrule(lr){2-4} \cmidrule(lr){5-7}
            & \textbf{AUC} $\uparrow$ & \textbf{Min L2} $\downarrow$ & \textbf{Avg L2} $\downarrow$  
            & \textbf{AUC} $\uparrow$ & \textbf{Min L2} $\downarrow$ & \textbf{Avg L2} $\downarrow$ \\
            \midrule
            Recasens et al.\cite{recasens2015they}  
            & - & - & -   
            & 0.878 & 0.113 & 0.19 \\
            Chong et al.\cite{chong2020detecting}  
            & - & - & -   
            & 0.921 & 0.077 & 0.137 \\
            Hu et al.\cite{hu2022gaze}  
            & - & - & -   
            & 0.923 & 0.069 & 0.128 \\
            Gupta et al.\cite{gupta2022modular}  
            & - & - & -   
            & 0.943 & 0.056 & 0.114 \\
            Tafasca et al.\cite{tafasca2024sharingan}  
            & 0.917 & 0.169 & 0.186   
            & 0.944 & 0.057 & 0.113 \\
            Gaze-LLE (B)\cite{ryan2024gaze}  
            & 0.918 & 0.190 & 0.205   
            & 0.956 & 0.045 & 0.104 \\
            Gaze-LLE (L)\cite{ryan2024gaze}  
            & 0.921 & 0.183 & 0.197   
            & \underline{0.958} & \underline{0.041} & \underline{0.099} \\
            \midrule
            \textbf{ANCHOR (B)} 
            & 0.928 & 0.181 & 0.197 
            & 0.955 & 0.052 & 0.112 \\
            \textbf{ANCHOR (L)} 
            &\textbf{0.932} & \textbf{0.173} &\textbf{0.188} 
            &\textbf{0.956} & \textbf{0.047} & \textbf{0.106} \\
            \bottomrule
            \end{tabular}%
        }
    \end{minipage}
    \hfill 
    \begin{minipage}[c]{0.38\linewidth}
        \centering
        \captionof{figure}{\textbf{UI of Social Influence Ranking Labeling.}}
        \includegraphics[width=\linewidth]{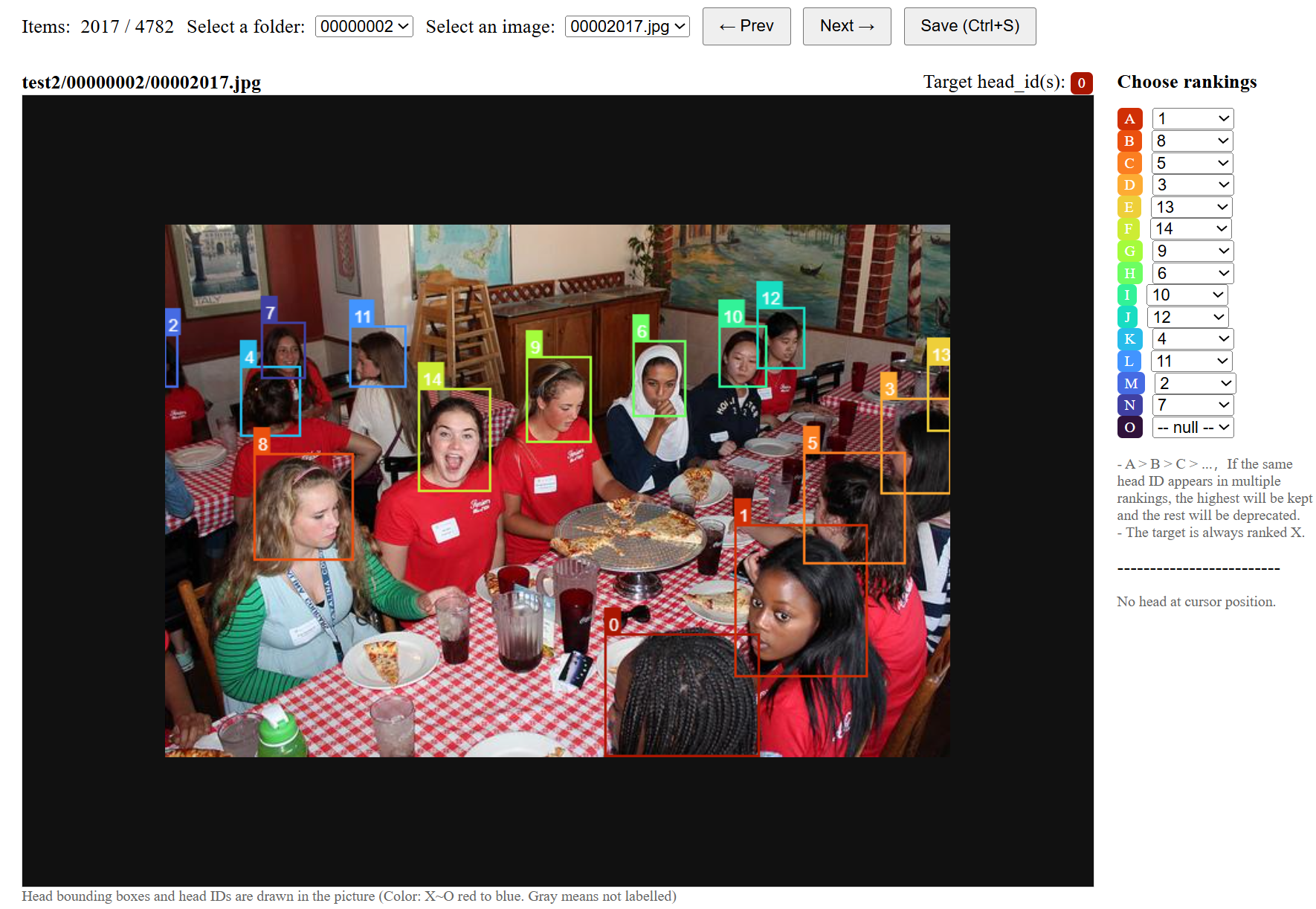}
        
        \label{fig:social_impact_UI}
    \end{minipage}
\end{table}

%% file: sec/4_method.tex
\section{Methodology}
\parahead{Problem Definition} Given an RGB image $\mathcal{I}$ and $K$ head bounding boxes $\mathbf{B}$, we aim to predict a person-specific gaze heatmap $\mathbf{H}_k \in [0, 1]^{H \times W}$ for every head $k$. Our premise is that accurate prediction requires modeling the joint distribution $P(\mathcal{H}, \mathbf{Z} | \mathcal{I}, \mathbf{B})$ by explicitly surfacing the latent social intent $\mathbf{Z}$ linking individual behaviors.
\input{fig_tex/head_distribution_3_DatasetsComp}

\subsection{Model Architecture}

ANCHOR(Fig.\ref{fig:framework}) integrates four primary components: a frozen backbone, a relational attention mechanism, a synergistic optimization strategy, and an evaluation protocol.

\parahead{Frozen Encoder} ANCHOR employs a frozen DINOv2 backbone  to extract shared scene features from image $\mathcal{I}$. Using the $K$ head bounding boxes in $\mathbf{B}$, we extract head-specific features $\mathbf{c}_k$ via masked average pooling over the corresponding spatial regions. These features are projected to a 256-dimensional space and, along with the scene tokens, processed by three stacked ViT blocks to yield refined global scene tokens $\mathbf{S}$.

\parahead{Relational Attention Mechanism}
Given the extracted features $\mathbf{c}_k$ and global tokens $\mathbf{S}$, ANCHOR decodes the latent social intent $\mathbf{Z}$ by first identifying the local social context $\mathbf{T}$ for a target head $k$. 
This context is formed by selecting the top-$N$ neighbors via a depth-adjusted distance $\ell_{adj} = \Vert \mathbf{p}_{k} - \mathbf{p}_{j} \Vert_{2} / \sqrt{A_{j}+\epsilon}$, where $\mathbf{p}_i$ denotes the normalized image-plane center of the $i_{th}$ head box and $A_j$ is the bounding box area. 
The area normalization incorporates monocular scale cues into image-plane proximity: for candidates with similar center distances, a larger head box yields a smaller adjusted distance and is therefore treated as a closer local neighbor.
We collect these distances into a geometric pseudo-label vector $\mathbf{D}_{inv} = [\ell_{adj}(k, j)^{-1}]_{j \in \mathcal{N}_k}$, representing the inverse proximity to neighbors.
We then employ a cross-attention block, using $\mathbf{c}_k$ as the query and $\mathbf{T}$ as the key-value pairs, to model the interplay between the individual and their surroundings. The output of this block, $\mathbf{X}_{social}^k$, represents social-aware features, while its internal attention weights $\mathbf{W}_{attn}$ serve as the explicit proxy for $\mathbf{Z}$. Finally, a FiLM module~\cite{perez2018film} modulates the scene tokens $\mathbf{X}_{scene}^k$ using parameters $(\gamma_k, \beta_k)$ derived from the social-aware features $\mathbf{X}_{social}^k$, yielding the socially-consistent feature map:
\begin{equation}
\mathbf{X}_{scene}^k = (1 + \gamma_{k}) \mathbf{X}_{social}^k + \beta_{k}
\end{equation}

\parahead{Joint Decoding}
ANCHOR yields a dual output for each target head: (1) the person-specific gaze heatmap $\mathbf{H}_k$, and (2) the latent social intent $\mathbf{Z}$. While $\mathbf{H}_k$ is produced by refining and upsampling the modulated features $\mathbf{X}_{scene}^k$ to $64 \times 64$ resolution, the social intent $\mathbf{Z}$ is explicitly represented by the attention weights $\mathbf{W}_{attn}$ extracted during the relational mechanism. This joint formulation enables the model to simultaneously recover spatial gaze targets and the underlying social net.

\subsection{Loss Design}
Our training objective is designed to maximize predictive accuracy for the primary gaze task while simultaneously enforcing the learning of latent social net in a stable manner.

\parahead{Target Gaze Loss $\mathcal{L}_{T}$}
The primary supervisory signal for the gaze task is the pixel-wise Binary Cross-Entropy (BCE) loss applied to the target head's heatmap prediction. The ground-truth heatmap is generated by placing a 2D Gaussian distribution (with $\sigma=3$) around the target gaze annotation. We define the target gaze loss $\mathcal{L}_{T}$ as the mean BCE over all target heads in the batch, $\mathcal{M}_{T}$:
\begin{equation}
\mathcal{L}_{T} = \frac{1}{|\mathcal{M}_{T}|} \sum_{k \in \mathcal{M}_{T}} \text{BCE}(\mathbf{H}_k, \mathbf{H}_{k}^{GT})  
\end{equation}

\parahead{Self-Supervised Social Consistency Loss $\mathcal{L}_{S}$}
To explicitly train the Social Cross-Attention block to decode the local social prior, we introduce the self-supervised Social Consistency Loss $\mathcal{L}_{S}$, which maximizes the correlation between the learned mean attention weights $\mathbf{W}_{attn}$ and the geometric pseudo-label $\mathbf{D}_{inv}$ defined in Sec. 4.1:
\begin{equation}
\mathcal{L}_{S} = - \text{Corr}(\mathbf{W}_{attn},\mathbf{D}_{inv})
\end{equation}
The final objective is $\mathcal{L}_{total} = \mathcal{L}_{T} + \lambda_{s} \mathcal{L}_{S}$. To mitigate potential gradient conflicts between these tasks, we employ the synergistic optimization strategy.

\subsection{Synergistic Optimization Strategy}

The joint modeling of gaze and social intent introduces a significant optimization challenge: gradient conflicts where $\nabla \mathcal{L}_{S}$ can degrade the primary gaze task performance. We resolve this via a two-stage optimization stack.

\parahead{Stage 1: Robustness via SAM}To ensure stable generalization and enforce a flat local minimum for the primary gaze task, we apply Sharpness-Aware Minimization (SAM)~\cite{foret2021sam} exclusively to the gaze objective $\mathcal{L}_T$. The robust gaze gradient $\mathbf{g}_{T}^{SAM}$ is derived by solving for the optimal neighborhood perturbation $\epsilon$:
\begin{equation}
\mathbf{g}_{T}^{SAM} = \nabla_{\theta} \mathcal{L}_{T}(\theta + \epsilon), \quad \text{where } \epsilon = \arg\max_{\|\epsilon\| \le \rho} \mathcal{L}_{T}(\theta + \epsilon)
\end{equation}

\parahead{Stage 2: Synergy via Gradient Projection}To mitigate interference, we utilize the Projecting Conflicting Gradients (PCGrad)~\cite{yu2020gradient} principle. We treat $\mathbf{g}_{T}^{SAM}$ as the anchor; if the social gradient $\mathbf{g}_{S}$ conflicts with it ($\mathbf{g}_{T}^{SAM} \cdot \mathbf{g}_{S} < 0$), $\mathbf{g}_{S}$ is projected orthogonally onto the normal plane of $\mathbf{g}_{T}^{SAM}$:
\begin{equation}
\mathbf{g}_{S}^{proj} = \mathbf{g}_{S} - \frac{\mathbf{g}_{T}^{SAM} \cdot \mathbf{g}_{S}}{\Vert \mathbf{g}_{T}^{SAM} \Vert_{2}^{2}} \mathbf{g}_{T}^{SAM}
\end{equation}
The final gradient update $\mathbf{g}_{final}$ combines the robust gaze anchor and the conflict-free social component, ensuring that relational learning only proceeds in directions that do not degrade gaze accuracy:
\begin{equation}
\mathbf{g}_{final} = \mathbf{g}_{T}^{SAM} + \lambda_{s} \cdot \mathbf{g}_{S}^{proj}
\end{equation}

%% file: fig_tex/head_distribution_3_DatasetsComp.tex
\begin{figure*}[t]
    \centering

    \begin{subfigure}[t]{0.48\linewidth}
        \centering
        \includegraphics[width=0.9\linewidth]{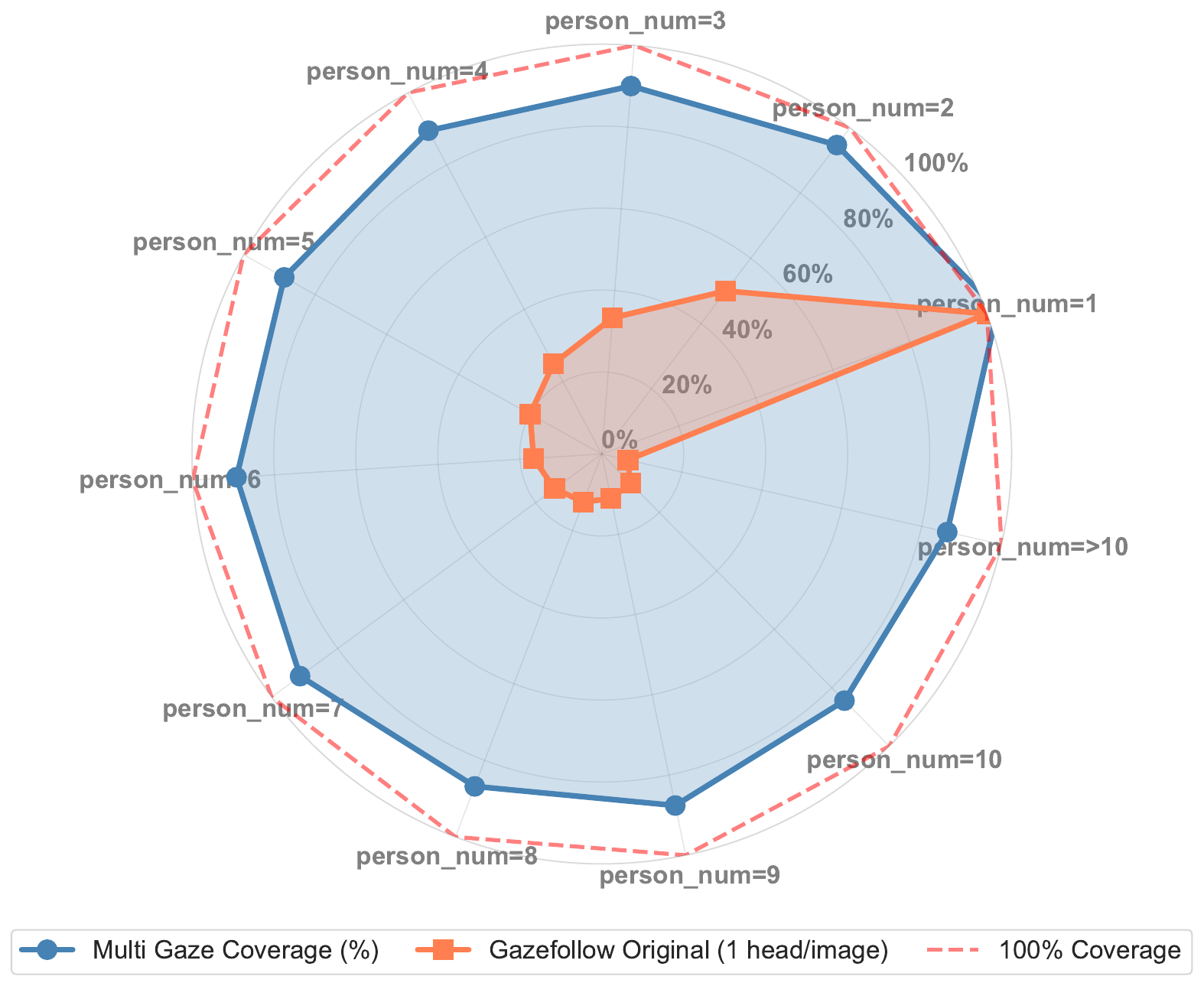}
        \caption{Multi-gaze label coverage.}
        \label{fig:head_number_hist}
    \end{subfigure}
    \hfill
    \begin{subfigure}[t]{0.48\linewidth}
        \centering
        \includegraphics[width=\linewidth]{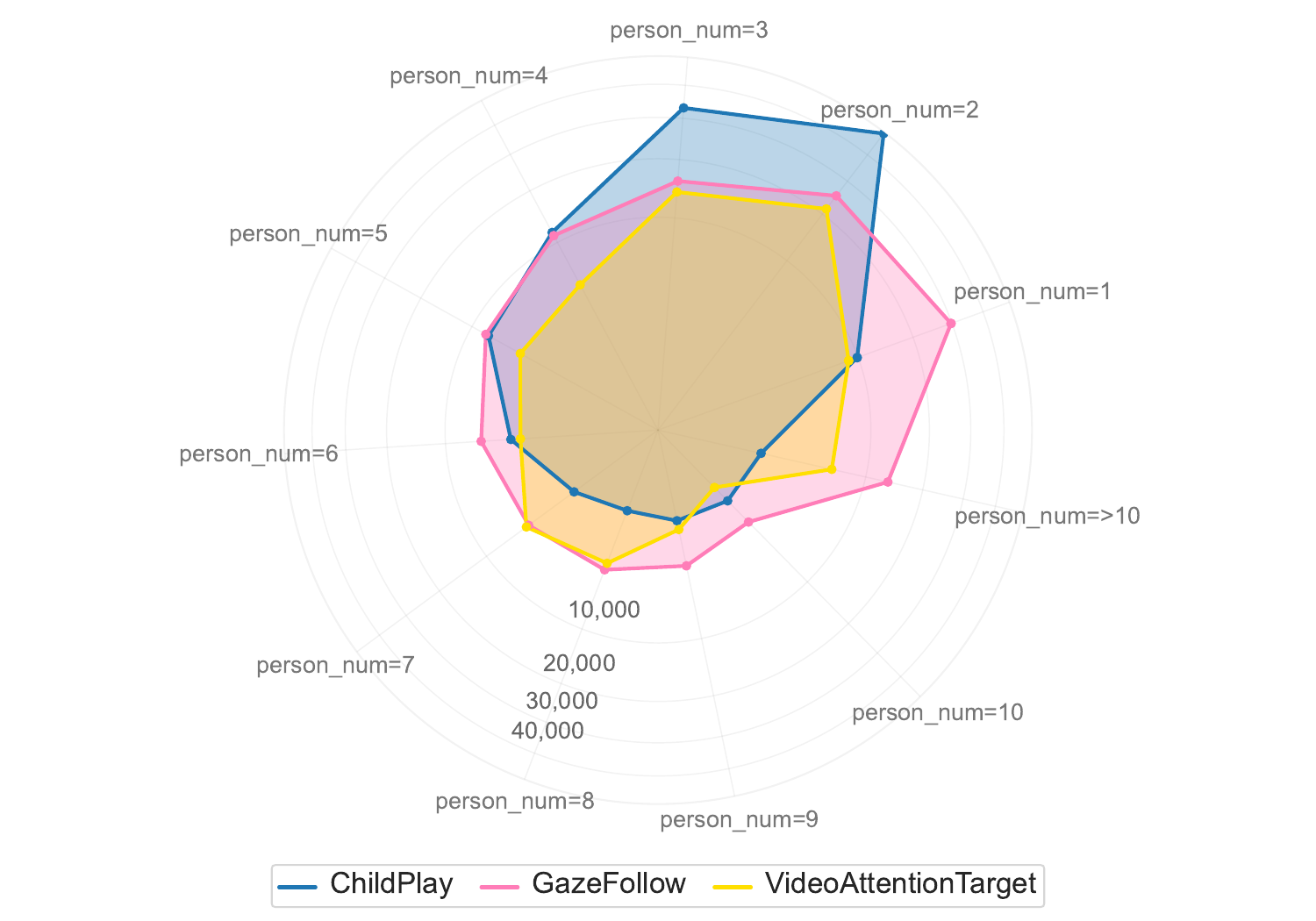}
        \caption{Datasets coverage comparison.}
        \label{fig:heads_distribution}
    \end{subfigure}

    \vspace{1.2em}

    \begin{minipage}[t]{0.4\linewidth}
        \begin{subfigure}[t]{\linewidth}
            \centering
            \vspace*{0pt}
            \resizebox{0.9\linewidth}{!}{
                \begin{tabular}{@{}lcc@{}}
                \toprule
                \textbf{Dataset} &
                \thead{\textbf{Multi-head}\\ \textbf{Gaze}} &
                \thead{\textbf{Social}\\ \textbf{Impact}} \\
                \midrule
                GazeFollow \cite{recasens2015they} & \xmark & \xmark \\
                VideoGaze \cite{recasens2016following} & \xmark & \xmark \\
                VideoCoAtt \cite{fan2018inferring} & \cmark & \xmark \\
                Gaze360 \cite{kellnhofer2019gaze360} & \cmark & \xmark \\
                VideoAttentionTarget \cite{chong2020detecting} & \cmark & \xmark \\
                GazeFollow360 \cite{Li_2021_ICCV} & \cmark & \xmark \\
                GOO \cite{tomas2021goo} & \xmark & \xmark \\
                ChildPlay \cite{tafasca2023childplay} & \cmark & \xmark \\
                \midrule
                \textbf{Ours (GF test ext.)} & \cmark & \cmark \\
                \bottomrule
                \end{tabular}
            }
            \caption{Comparison of existing gaze datasets. 
            }
            \label{tab:datasets}
        \end{subfigure}
    \end{minipage}
    \hfill
    \begin{minipage}[t]{0.48\linewidth}
        \begin{subfigure}[t]{\linewidth}
            \centering
            \vspace*{0pt}
            \includegraphics[width=\linewidth]{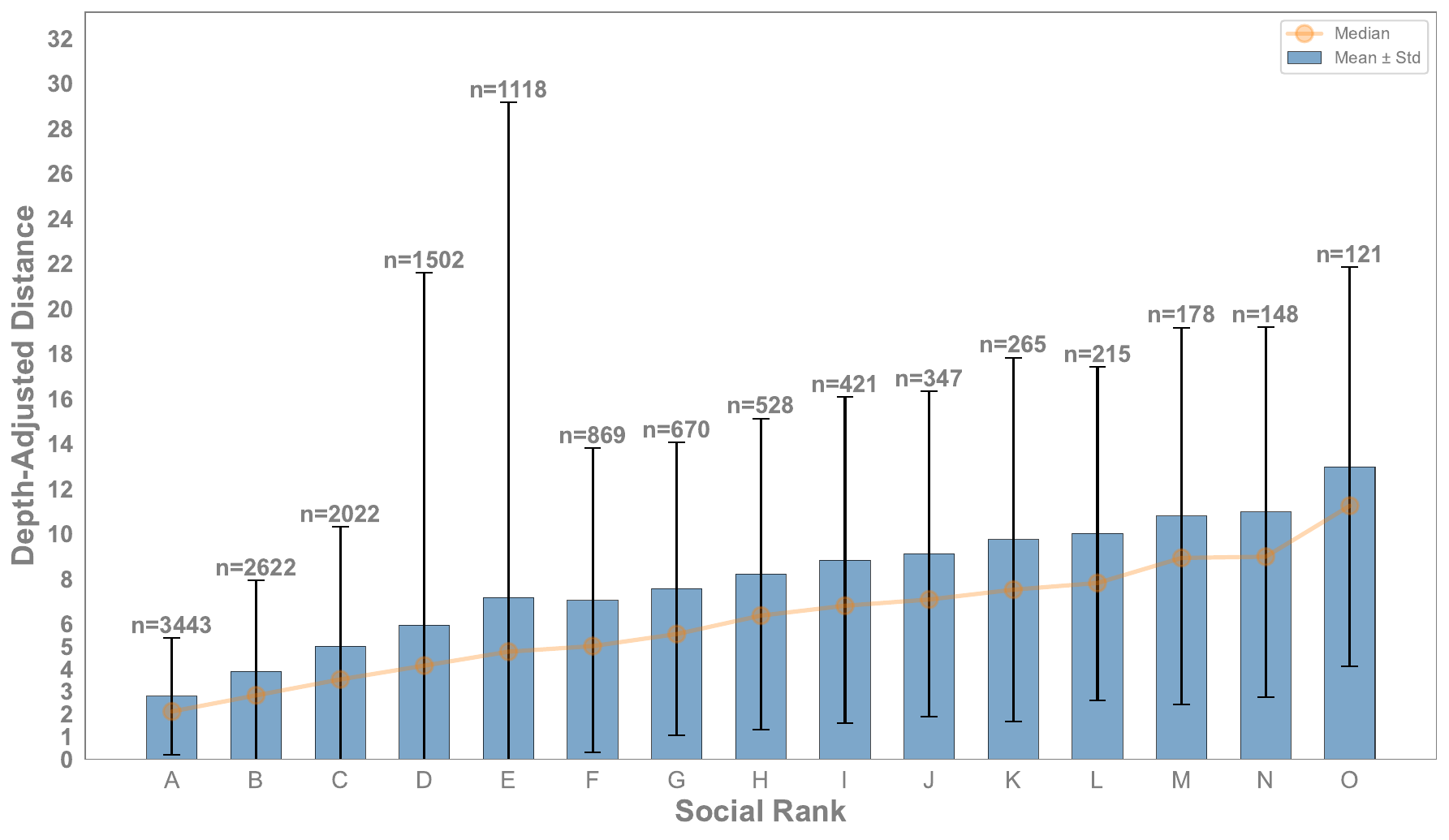}
            \caption{Distances across social ranks.}
            \label{fig:rankings_distance}
        \end{subfigure}
    \end{minipage}

    \caption{
        \textbf{Dataset Characteristics:}
        (a) Gaze annotation coverage across head-number groups.
        (b) Coverage comparison of existing multi-person gaze datasets.
        (c) Comparison of dataset annotation properties with prior work.
        Ours is the extended GazeFollow test split.
        (d) Average depth-adjusted distances across social ranks ($\pm\sigma$).}
    
\label{fig:combined_all}
\end{figure*}

%% file: sec/5_exp.tex
\section{Experiments}
\label{sec:experiments}
Our evaluation focus on three aspects: (i) the target-centric architecture’s superiority over global-map models, (ii) the synergistic optimization strategy in resolving task-level gradient conflicts, and (iii) the quantitative decoding of latent social intent. Results demonstrate that ANCHOR achieves state-of-the-art static multi-gaze accuracy while successfully recovering underlying social net.

\input{fig_tex/social_intent1}

\subsection{Evaluation Metrics}
\label{sec:evaluation_metrics}
To validate the framework's performance, we use standard gaze metrics and introduce novel social evaluation metrics enabled by our test set extensions.

\parahead{Gaze Prediction Metrics}
We evaluate gaze prediction accuracy using three metrics: AUC (Area Under the Curve), Min L2 Distance (minimum Euclidean distance between predicted maximum and all ground-truth points), and Avg L2 Distance (Euclidean distance between predicted maximum and the mean ground-truth point). For all metrics, we report performance in two scopes: the Target Gaze metric, which aligns with the standard GazeFollow benchmark by focusing only on the single designated person per image, and the comprehensive Multi Gaze metric ('\_multi' suffix) validating the joint-distributional model across all annotated heads in the scene.

\parahead{Integrated Social-Relational Metrics} 
To quantify the recovery of both geometric priors and perceived social ranking, we introduce three complementary metrics: Social Consistency Index (SCI), Social Leader Accuracy (SLA), and Social Influence Rank Correlation ($\rho_{SIR}$). While the observed negative correlation between social impact and distance $l_{adj}$ (Fig.~\ref{fig:rankings_distance}) justifies our $\mathcal{L}_{SCI}$ prior, high variance indicates that social intent is a complex latent variable exceeding mere geometry. This underscores the necessity of our multi-faceted evaluation—using SCI to validate internal geometric mechanisms, SLA to measure practical leader recognition, and $\rho_{SIR}$ to assess the alignment with the full human-annotated hierarchy. Metrics are calculated per target head $k$ and averaged over the dataset.

\parahead{\textit{1. Social Consistency Index (SCI)}}This index validates the relational mechanism by measuring how attention weights $W_{attn}$ align with geometric proximity priors. For a target $k$, we compute the Pearson correlation~\cite{pearson1895notes} between the attention vector $\mathbf{w}^{(k)}$ and the inverse distance vector $\mathbf{d}_{\text{inv}}^{(k)} = [\ell_{\text{adj}}(k,n)^{-1}]_{n \in \mathcal{N}_k}$:
\begin{equation}\text{SCI}^{(k)} = \text{Corr}_{\text{Pearson}}\left(\mathbf{w}^{(k)},\mathbf{d}_{\text{inv}}^{(k)}\right).\label{eq:sci}
\end{equation}
A high SCI indicates that the model successfully learns the linear proportionality hypothesis ($W_{\text{attn}} \propto \ell_{\text{adj}}^{-1}$) defined in our self-supervised objective.

\parahead{\textit{2. Social Leader Accuracy (SLA)}}This metric assesses the model's ability to identify the most salient individual (the "leader") in a social scene. We define the prediction as correct if the neighbor $n^*$ receiving the maximum attention weight belongs to the highest human-annotated social ranking $\mathcal{R}_{gt}{top1}$:
\begin{equation}
\text{SLA} = \frac{1}{|\mathcal{M}|} \sum_{k \in \mathcal{M}} \mathbb{I}\left( R\left(\text{argmax}_{j \in \mathcal{N}k} w_j^{(k)}\right) \in \mathcal{R}_{gt}^{top1} \right).
\label{eq:sla}
\end{equation}
SLA provides a discrete validation of the model's practical social reasoning capability in identifying centers of attention.

\parahead{\textit{3. Social Influence Rank Correlation ($\rho_{\text{SIR}}$)}}To evaluate whether ANCHOR captures the full human-perceived social hierarchy, we use Spearman’s~\cite{spearman1904proof} rank correlation. This compares the model-derived ordering of neighbors against the human-annotated ordinal ranks $\mathcal{R}_{\text{gt}}^{(k)}$:
\begin{equation}\rho_{\text{SIR}}^{(k)} = \text{Corr}_{\text{Spearman}}\left(\mathcal{R}(\mathbf{w}^{(k)}), \mathcal{R}_{\text{gt}}^{(k)}\right).\label{eq:sir}
\end{equation}
Unlike social leader accuracy, $\rho_{\text{SIR}}$ is invariant to scale and measures the monotonic agreement across the entire social graph, confirming the model's depth in decoding complex relations beyond simple geometry.

\subsection{Implementation Details}
We train on the standard GazeFollow dataset~\cite{recasens2015they} and evaluate on our Extended GazeFollow Benchmark. All models use a frozen DINOv2 backbone and a consistent latent dimension of $C=256$. Our final model is trained for 15 epochs. 

\subsection{Qualitative Analysis}
Fig.~\ref{fig:qualitative_combined} illustrates ANCHOR coupling gaze prediction with target-aware social reasoning. In the first row, multi-gaze heatmaps align with the influence graph, where shared targets and proximity correlate with higher ranks and thicker connections. The model moves beyond simple distance priors, leveraging joint gaze direction and body orientation to assign social influence. The second row shows ANCHOR correctly recovering top influential neighbours (ranking A, B, C) even among equidistant candidates. While minor discrepancies occur in lower ranks where human labels are less sharp, re-targeting rows demonstrate plausible graph reorientation toward new focal subjects. These results confirm a flexible, target-centric social representation that successfully captures the latent structural scaffolding of gaze behavior.

\subsection{Quantitative Analysis}
\subsubsection{Comparison to State-of-art Methods}

We compare ANCHOR with state-of-the-art methods on two tasks(Tab~\ref{tab:gazefollow_multi}): (1) the full Multi Gaze task, which evaluates true joint-distributional performance, and (2) the standard single-person Target Gaze task. On Multi Gaze (Table~\ref{tab:gazefollow_multi}), ANCHOR (DINOV2-L) achieves an AUC of 0.932, Min L2 of 0.173, Avg L2 of 0.188, surpassing the best i.i.d. model, Gaze-LLE (L) (AUC=0.921,  Min L2=0.183, Avg L2=0.197). This demonstrates the empirical cost of the conditional independence assumption.
The marginal gap between ANCHOR and Gaze-LLE on Target Gaze is expected. While Gaze-LLE focuses solely on i.i.d. pixel-level accuracy, ANCHOR is designed to decouple high-quality social latent variables from complex scene contexts. This necessitates a trade-off: we accept a minimal sacrifice in individual AUC to achieve jointly-optimized, socially-consistent predictions, effectively trading narrow task-specific precision for meaningful social-relational understanding.

\input{tab/2_model_ablation}

\subsubsection{Ablation Study}
We conduct a series of ablations to validate our two primary claims: the necessity of the ANCHOR architecture and the  synergistic optimization stack.

\noindent \textbf{Backbone Selection Analysis:}
We summarize here only the key outcome of our backbone study. Frozen DINOv2 backbones consistently provide the most stable and accurate performance, while partial or full fine-tuning leads to severe degradation. Detailed results, comparisons with DINOv3, and the full backbone selection analysis are provided in the supplementary material.

\noindent \textbf{Component Ablation Analysis:}
Table~\ref{tab:model_ablation} reveals the progressive emergence of joint-distributional capability in our model. The frozen DINOv2 baseline provides strong single-head performance but fails to capture multi-person consistency, indicating that backbone strength alone is insufficient. Simply adding FiLM or the Social Cross-Attention block does not yield meaningful gains and can even degrade results, demonstrating that naive architectural modifications cannot resolve the core challenge. Substantial improvement first appears when introducing the self-supervised $\mathcal{L}_{SCI}$ objective, which encourages the model to internalize geometric-social relations and raises multi-head accuracy accordingly. The decisive transition, however, comes from the full  synergistic optimization strategy, which stabilizes optimization and prevents the collapse observed in earlier variants, enabling the model to jointly optimize social reasoning and gaze localization. Scaling the backbone to ViT-L further strengthens performance. Overall, the ablation shows that $\mathcal{L}_{SCI}$ provides the necessary supervisory signal for extracting latent social structure, while  synergistic optimization strategy is essential for reliably integrating these signals within the shared representation space, making both components indispensable to the final ANCHOR architecture.

\subsection{Qualitative and Social Analysis}

Finally, we analyze our model's performance in complex scenes and its ability to decode social intent.

\noindent \textbf{Robustness in Complex Scenes:}
Figure~\ref{fig:model_comparison_3Model_multiHeadNum} demonstrates ANCHOR's robustness. Our model (ViT-L, yellow) consistently outperforms the SOTA i.i.d. model Gaze-LLE (ViT-L, gray) on Multi Gaze AUC across all head counts. The performance gap \textit{widens} in high-density scenes (e.g., 8, 10, and 15 heads), proving our joint-distributional model is superior at handling the complex dependencies that i.i.d. models ignore.
\input{fig_tex/model_comparison_3Model_multiHeadNum}

\noindent \textbf{Quantitative Proof of Social Intent:}
We evaluate the decoded social vector $\mathbf{W}_{\text{attn}}$ using our three social metrics.
\textbf{SCI} provides mechanism-level validation: a positive mean score ($0.282$) shows that the SCA block successfully learns the geometric prior encouraged by $\mathcal{L}_{SCI}$.  
\textbf{SLA} (Top-1 High-Level Accuracy) offers the clearest practical evidence of decoded intent: ANCHOR (ViT-L) achieves an \textbf{88.2\%} accuracy in identifying the highest-tier social neighbor, demonstrating strong leader recognition.  
Finally, \textbf{$\rho_{\text{SIR}}$} captures the full social hierarchy; a significant positive correlation ($0.280$) shows the model recovers not only the leader but the continuous ranking structure perceived by humans.

%% file: fig_tex/social_intent1.tex
\begin{figure*}[t] %
    \centering
\begin{minipage}[b]{\linewidth}
    \centering
    \begin{subfigure}[b]{0.48\linewidth}
        \centering
        \includegraphics[width=\linewidth]{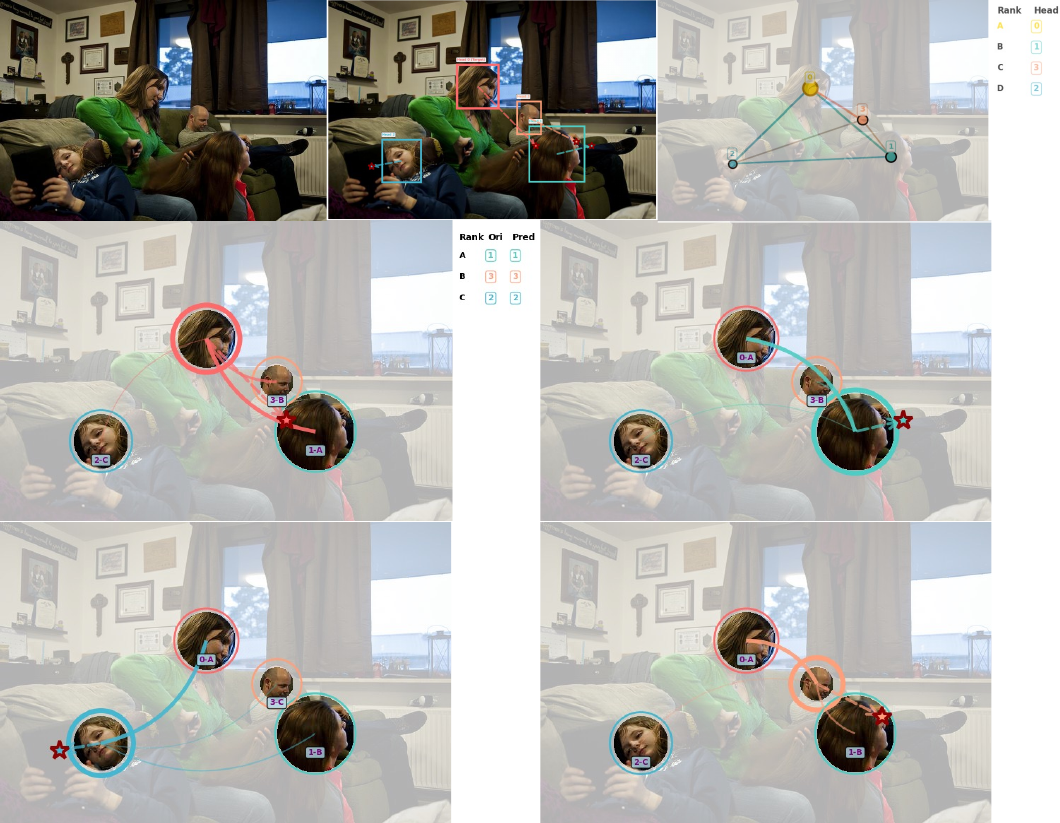}
        \caption{} 
        \label{fig:social_intent4}
    \end{subfigure}
    \hfill 
    \begin{subfigure}[b]{0.38\linewidth}
        \centering
        \includegraphics[width=\linewidth]{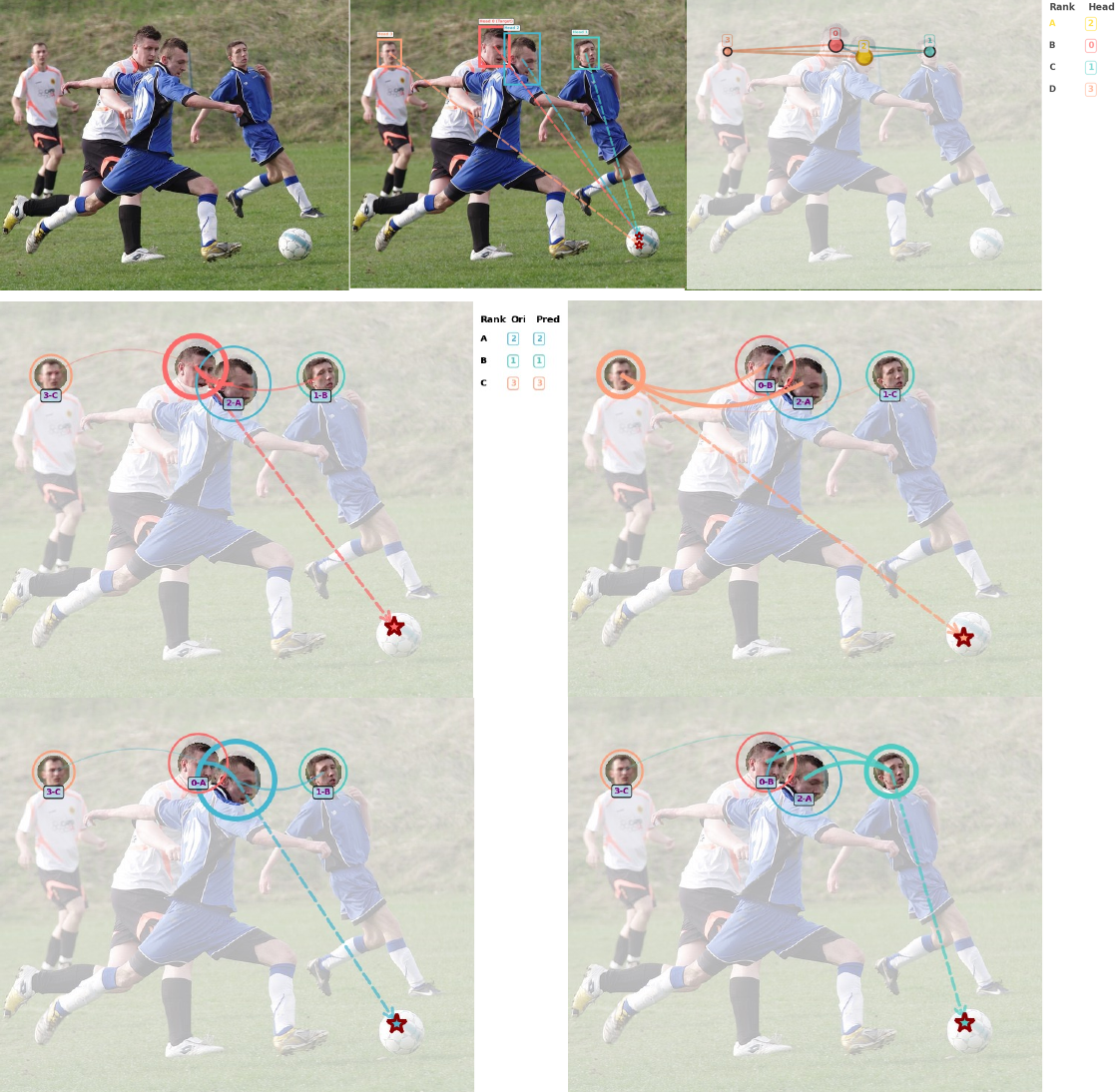}
        \caption{} 
        \label{fig:social_intent3}
    \end{subfigure}
\end{minipage}

    \caption{\textbf{Visualisation of multi-person gaze prediction and decoded social net.}}
    \label{fig:qualitative_combined}
\end{figure*}

%% file: tab/2_model_ablation.tex
\begin{table*}[t]
\centering
\caption{\textbf{Model ablation study.} Comparison of different model variants on \textbf{GazeFollow}. 
Multi Gaze columns measure joint consistency across people, while Single Gaze columns evaluate individual prediction quality.}
\resizebox{0.99\linewidth}{!}{
\begin{tabular}{l | c c c c | c c c | c c c}
\toprule
\multirow{2}{*}{\textbf{Backbone}} &
\multicolumn{4}{c|}{\textbf{Ablation Configuration}} &
\multicolumn{3}{c|}{\textbf{Multi Gaze}} &
\multicolumn{3}{c}{\textbf{Single Gaze}} \\
\cmidrule(lr){2-5}\cmidrule(lr){6-8}\cmidrule(lr){9-11}
& w/o FiLM & w/o Cross Attn  & w/o $L_{\text{SCI}}$ & w/o TA-RO &
AUC $\uparrow$ & Min L2 $\downarrow$ & Avg L2 $\downarrow$ &
AUC $\uparrow$ & Min L2 $\downarrow$ & Avg L2 $\downarrow$ \\
\midrule
DINOv2(B) & \xmark & \xmark & \xmark & \xmark &
0.916 & 0.193 & 0.206 &
0.955 & 0.048 & 0.108 \\
DINOv2(B) & \cmark & \xmark  & \xmark & \xmark &
0.916 & 0.215 & 0.228 &
0.949 & 0.050 & 0.109 \\
DINOv2(B) & \cmark & \cmark  & \xmark & \xmark &
0.916 & 0.215 & 0.228 &
0.949 & 0.051 & 0.109 \\
DINOv2(B) & \cmark & \cmark  & \cmark & \xmark &
0.920 & 0.214 & 0.226 &
0.951 & 0.049 & 0.108 \\
DINOv3 (B) & \cmark & \cmark & \cmark & \cmark 
& 0.915 & 0.223 & 0.235 & 
0.951 & 0.051 & 0.110 \\
DINOv3 (L) & \cmark & \cmark  & \cmark & \cmark & 
0.894 & 0.246 & 0.263 & 
0.932 & 0.096 & 0.162 \\
ViT(S) & \cmark & \cmark  & \cmark & \cmark & 
 0.852 & 0.313  & 0.328 & 
 0.887 & 0.150 & 0.222  \\
CNN(ResNet50) & \cmark & \cmark  & \cmark & \cmark & 
0.830  & 0.329  & 0.344  & 
0.876  & 0.156  & 0.230  \\
\midrule
\textbf{DINOv2(B)} &
\textbf{\cmark} & \textbf{\cmark}  & \textbf{\cmark} & \textbf{\cmark} &
0.928 & 0.181 & 0.197 &
0.955 & 0.052 & 0.112 \\
\textbf{DINOv2(L)} &
\textbf{\cmark} & \textbf{\cmark} & \textbf{\cmark} & \textbf{\cmark} &
\textbf{0.932} & \textbf{0.173} &\textbf{0.188} &
\textbf{0.956} & \textbf{0.047} & \textbf{0.106} \\
\bottomrule
\end{tabular}
}
\label{tab:model_ablation}
\end{table*}

%% file: fig_tex/model_comparison_3Model_multiHeadNum.tex




\begin{figure*}[t]
    \centering

    \begin{subfigure}[b]{1\linewidth}
        \centering
        \includegraphics[width=1\linewidth]{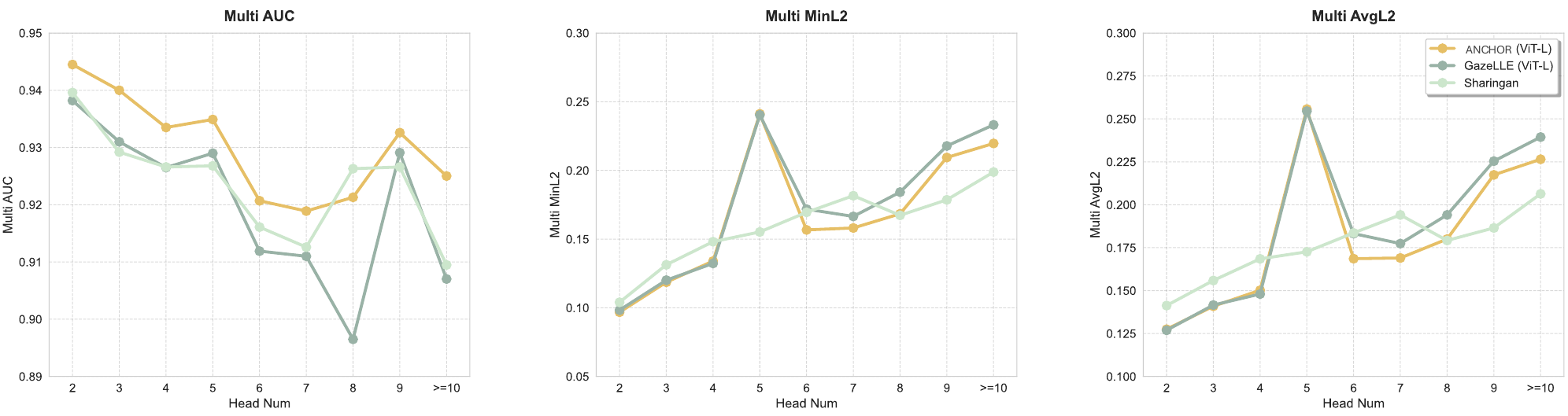}
        \caption{} 
        \label{fig:model_comparison_3Model_multiHeadNum}
    \end{subfigure}

    \vspace{0.1em} 

    \begin{subfigure}[b]{0.48\linewidth}
        \centering
        \includegraphics[width=\linewidth]{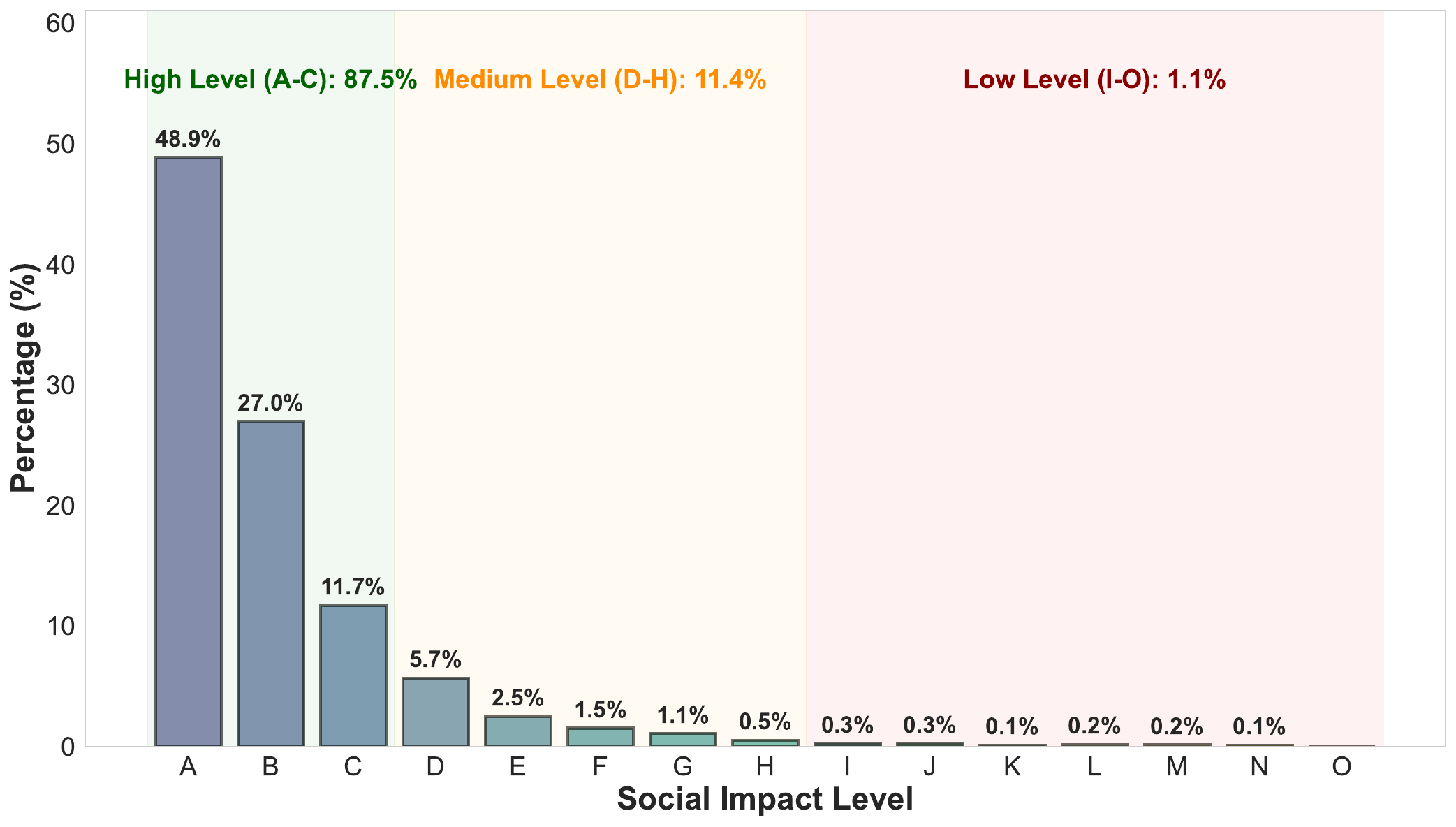}
        \caption{} 
        \label{fig:top1_acc}
    \end{subfigure}
    \hfill 
    \begin{subfigure}[b]{0.48\linewidth}
        \centering
        \includegraphics[width=\linewidth]{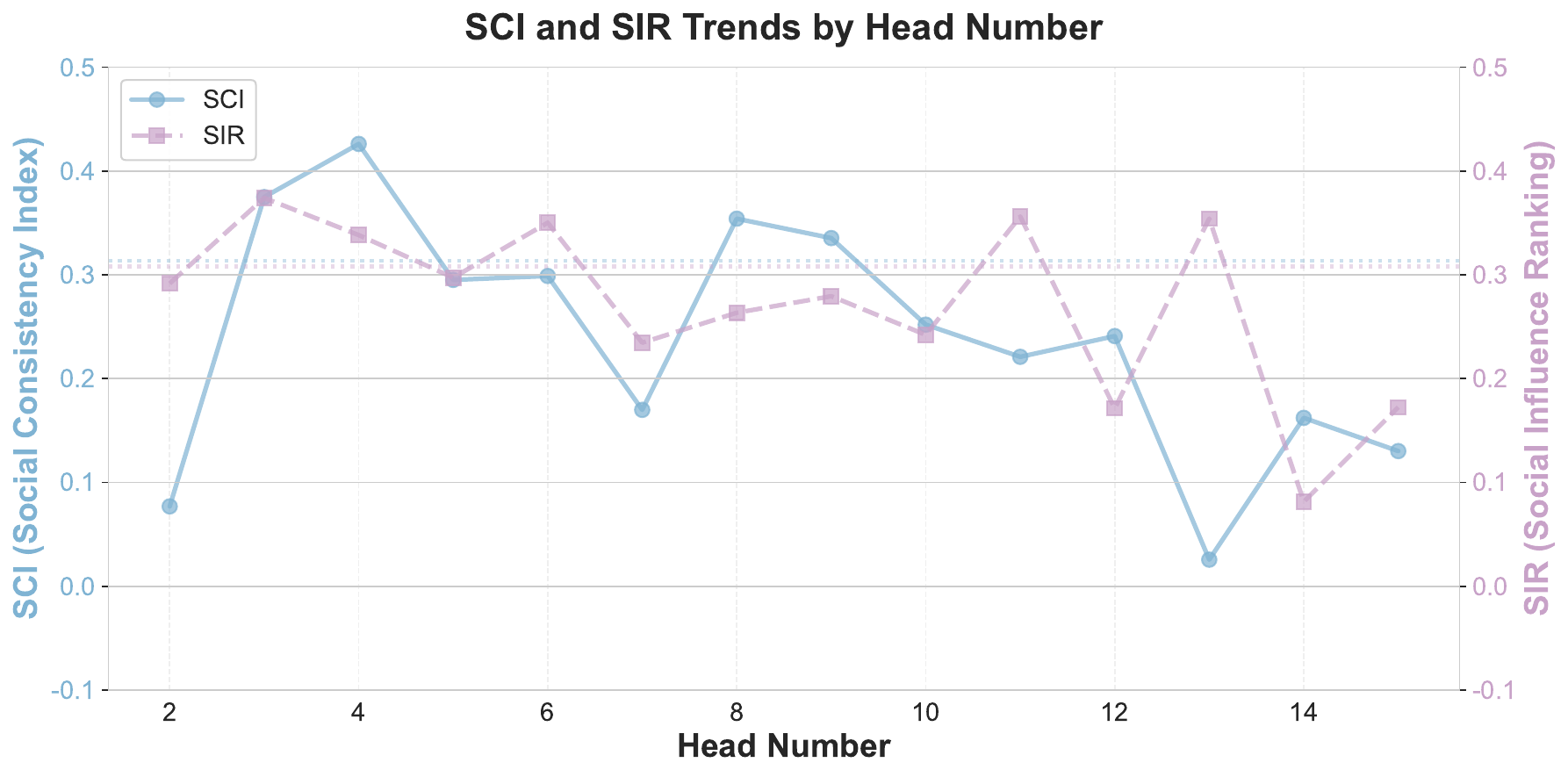}
        \caption{} 
        \label{fig:sci_sir_dual_axis}
    \end{subfigure}

    \caption{\textbf{Comprehensive Model Performance and Social Analysis:} 
        (a) Gaze following performance comparison (AUC and $L_2$ distance) across varying head numbers for ViT-B and ViT-L backbones.
        (b) Detailed analysis of Social Leader Accuracy (SLA) for Top-1 High-Level neighbor prediction across different social tiers.
        (c) Measurements of Social Consistency Index (SCI) and Social Influence Ranking ($\rho_{\text{SIR}}$) metrics across impact levels.}
    \label{fig:combined_all_results}
\end{figure*}

%% file: sec/6_discussion.tex
\section{Discussion}
\label{sec:discussion}
Our experiments validate ANCHOR’s key claims. First, single-target evaluation hides the challenge of joint gaze reasoning: i.i.d.-optimized baselines such as Gaze-LLE perform well on Single Gaze but drop on Multi Gaze, whereas ANCHOR (ViT-L) reaches 0.932 AUC and stays robust in complex scenes. Second, ablations show FiLM and SCA offer little alone; gains arise only with $\mathcal{L}_{SCI}$, and stable performance requires TA-RO, confirming that exposing $X_{social}$ creates a gradient conflict needing robust, projection-aware optimization. Third, our social metrics validate decoded intent: SCI captures geometric priors, SLA (88.2\%) finds the most influential neighbor, and $\rho_{SIR}$ aligns with human rankings. These results provide the first quantitative evidence that models can infer latent social structure from gaze cues. Remaining limitations include dependence on a frozen backbone, tunable optimization weights, and the focus on static images, suggesting future extension to temporal settings.

%% file: sec/7_conclusion.tex
\section{Conclusion}

In this work, we address the conditional-independence flaw in gaze-following research by modeling the joint distribution of gaze and latent social intent. We introduce ANCHOR, which integrates four primary components: (1) a frozen backbone for robust feature extraction; (2) a relational attention mechanism to decode social context; (3) a synergistic optimization strategy to resolve gradient conflicts; and (4) a valuation protocol featuring the Extended GazeFollow Benchmark. Our experiments demonstrate that ANCHOR achieves state-of-the-art multi-head performance while proving that the synergistic optimization strategy is essential for stable training. Furthermore, through our SCI, SLA and $\rho_{SIR}$ metrics, we provide the first quantitative evidence that implicit social dynamics can be reliably decoded from static gaze patterns.